\documentclass[lettersize,journal]{IEEEtran}
\usepackage{amsmath,amsfonts}
\usepackage{algorithmic}
\usepackage{algorithm}
\usepackage{array}
\usepackage[caption=false,font=normalsize,labelfont=sf,textfont=sf]{subfig}
\usepackage{textcomp}
\usepackage{stfloats}
\usepackage{url}
\usepackage{verbatim}
\usepackage{graphicx}
\usepackage{cite}
\usepackage{booktabs}
\usepackage{adjustbox}
\usepackage{hyperref}
\usepackage{multirow} 
\usepackage{xcolor}
\newcommand{\rev}[1]{\textcolor{black}{#1}}
\newcommand{\revred}[1]{\textcolor{black}{#1}}
\newcommand{\revorange}[1]{\textcolor{black}{#1}}

\begin{document}

\title{Neural-Collapse-guided Task-Free Continual Anomaly Detection}

\author{
Xiaotong Kong, Chaoyang Song, Ziai Zhou, Jinxia Zhang*, Kanjian Zhang, Haikun Wei%
\thanks{* Corresponding author. 
This work was supported by National Natural Science Fund of China (Grant number 62573123), Research Fund for Advanced Ocean Institute of Southeast University, Nantong (GP202411), the Postgraduate Research \& Practice Innovation Program of Jiangsu Province (26CXJH0649) and the Fundamental Research Funds for the Central Universities. We also thank the Big Data Computing Center of Southeast University for providing the facility support on the numerical calculations in this paper.

Xiaotong Kong, Chaoyang Song, Ziai Zhou, Jinxia Zhang, Kanjian Zhang and Haikun Wei are with the Key Laboratory of Measurement and Control of CSE, Ministry of Education, School of Automation, Southeast University, Nanjing 210096, China (e-mail: \{230259964, 230248980, 230238499, jinxiazhang, kjzhang, hkwei\}@seu.edu.cn).
}
}

\maketitle

\begin{abstract}
Recent years have witnessed growing interest in continual anomaly detection for industrial visual inspection. However, real-world manufacturing environments exhibit unpredictable shifts in data distributions, rendering task-dependent continual learning assumptions impractical. \revorange{To address this limitation, we formulate industrial anomaly detection as a task-free continual learning problem and propose NC-TFAD, a neural-collapse-inspired, geometry-driven framework for learning from non-stationary data streams without task boundaries.} NC-TFAD freezes a pretrained backbone and aligns streaming features to a simplex Equiangular Tight Frame (ETF) prototype space to stabilize representation geometry under non-stationary streams. To satisfy the NC-inspired geometric construction in the absence of real anomalies, we generate synthetic anomaly samples as auxiliary anchors during training. Building on this geometry, we further introduce inter- and intra-class regularization together with a Focal Neural Collapse Contrastive (FNCC) loss to suppress representation drift and improve normal–anomaly separability. Finally, \revorange{a normal-patch-prototype-guided localization branch constructs calibrated patch-wise deviation maps from normal training samples and fuses them with a weak self-attention prior, producing anomaly heatmaps without pixel-level annotations}. Extensive experiments on MVTec AD and VisA show that NC-TFAD consistently outperforms representative task-free continual learning methods adapted from general vision, as well as unified anomaly detection baselines, in both image-level detection and pixel-level localization \revorange{under the task-free continual learning protocol}. These results highlight that geometry-driven modeling offers an effective and robust solution for task-free continual anomaly detection in real-world industrial applications. 
\end{abstract}

\begin{IEEEkeywords}
Anomaly detection, Task-free, Continual learning, Neural collapse
\end{IEEEkeywords}

\section{Introduction}
\IEEEPARstart{A}{nomaly} detection (AD) is a fundamental problem in industrial visual inspection and intelligent manufacturing systems. It is critical for ensuring product quality, reducing production risks, and enabling automated workflows \cite{ref1}, \cite{ref2}. In practice, anomalies exhibit diverse appearances, wide scale variation, and severe class imbalance. These factors make manual, expertise-driven inspection inefficient and inconsistent in accuracy and repeatability. As industrial data volumes grow, manual inspection can no longer meet the requirements of high precision, low latency, and long-term reliability, accelerating the adoption of deep-learning-based AD in industrial settings.

In recent years, deep learning has significantly advanced the development of anomaly detection methods \cite{ref3}, \cite{ref4}. Compared with rule-based or shallow-feature approaches, deep learning methods can learn representations directly from data. They can exploit large collections of normal samples to acquire discriminative features. As a result, they typically generalize better under complex backgrounds, diverse defect types, and cross-category scenarios. Existing deep anomaly detection methods have evolved from the paradigm of one-model-one-category toward one-model-for-all \cite{ref5},\cite{ref6}. Earlier category-specific models perform well in controlled settings but require frequent retraining as products or production lines change, leading to high maintenance cost and poor scalability. Unified frameworks such as UniAD\cite{ref7} alleviate these issues by consolidating multiple categories into a single model and reducing deployment and storage overhead.

Since most existing methods assume that all training data are jointly available during model training \cite{ref1}, \cite{ref8}, \cite{ref9}, \cite{ref10}, a substantial gap remains between unified AD models and real industrial deployment. In practical manufacturing environments, new product categories and inspection objects are introduced incrementally. In this context, continual learning (CL) \cite{ref11}, \cite{ref12} has been adopted in anomaly detection to address distributional shifts over time. Nevertheless, conventional CL methods generally rely on explicit task boundaries or task identifiers. Such information is rarely available in industrial AD scenarios, where task transitions are inherently ambiguous and task labels are unavailable during inference\cite{ref13}. Consequently, task-free continual learning (TFCL) is regarded as a more realistic and deployment-oriented learning paradigm for industrial anomaly detection.

While TFCL has achieved promising progress in general vision tasks \cite{ref14}, \cite{ref15}, its application to industrial anomaly detection remains limited. Existing task-free frameworks are not specifically designed for anomaly detection and struggle to reconcile long-term knowledge retention with the detection of subtle anomalies. The problem is further exacerbated by the lack of anomaly supervision.

To bridge this gap, we formulate industrial AD as a task-free continual learning problem and propose Neural-Collapse-guided Task-Free Anomaly Detection (NC-TFAD), a geometry-driven framework specifically tailored for this demanding setting. Unlike existing continual or unified anomaly detection approaches that rely on explicit task boundaries, replay buffers, or joint training assumptions, NC-TFAD enables continual adaptation under unknown task transitions using only the currently observed data stream. Specifically, NC-TFAD freezes a pretrained backbone to extract stable visual features and employs a lightweight  linear projection layer to align streaming features to a Simplex Equiangular Tight Frame (ETF) prototype space induced by neural collapse\cite{ref16}.To instantiate the NC-inspired normal--anomaly geometry in the absence of real anomaly supervision, 
\revred{we introduce a synthetic anomaly generation strategy that provides auxiliary geometric anchors for the otherwise unobserved anomalous direction. The resulting synthetic anomalies provide weak image-level supervision for establishing a stable normal--anomaly geometric reference, rather than explicitly modeling the full appearance distribution of real industrial defects.}
Building on this geometric structure, we introduce NC-guided inter- and intra-class regularization to preserve the ETF-induced geometry by promoting intra-class compactness and inter-class separation, thereby stabilizing the embedding space under distribution shifts. We further propose a Focal Neural Collapse Contrastive (FNCC) loss that emphasizes hard sample--sample and sample--prototype pairs, improving sample-level discriminability and anomaly separability under non-stationary streams. 
\revorange{In addition, NC-TFAD includes a normal-patch-prototype-guided localization branch. It builds a compact prototype set from normal training patches only, scores each test patch by its calibrated deviation from the normal patch manifold, and uses backbone self-attention only as a weak spatial prior.}

Extensive experiments on the MVTec AD and VisA datasets demonstrate that NC-TFAD significantly outperforms existing state-of-the-art approaches under the task-free continual anomaly detection setting.

The main contributions of this work are summarized as follows:
\begin{enumerate}

\item\revorange{We formulate industrial anomaly detection under a task-free continual learning protocol in which normal samples arrive in a single-pass non-stationary stream without task boundaries, category identifiers, or access to historical data, better reflecting real-world industrial inspection with continuously evolving data distributions.}

\item \revred{We propose NC-TFAD, which reformulates neural-collapse geometry for task-free continual anomaly detection in a category-agnostic manner. Unlike standard multiclass ETF formulations tied to semantic classes, NC-TFAD employs two fixed ETF directions to represent normal and anomalous states shared across all categories. Since real anomalies are unavailable during training, synthetic anomalies are introduced as auxiliary geometric anchors to instantiate the missing anomalous direction without anomaly supervision.}

\item \revred{We develop an NC-guided continual optimization mechanism that integrates inter-class alignment and intra-class compactness with a Focal Neural Collapse Contrastive (FNCC) loss. FNCC jointly models sample--sample and sample--prototype relations within a unified contrastive objective and applies focal weighting to hard positive pairs and ambiguous prototype assignments, improving robustness under non-stationary distributions.}

\item\revorange{We further introduce a normal-patch-prototype-guided localization branch that uses only normal training patches to construct a compact localization memory. During inference, patch-wise prototype deviations are calibrated by the normal distance distribution and fused with a weak self-attention prior, improving pixel-level localization without requiring pixel-level supervision.}

\end{enumerate}

\section{Related Works}
\subsection{Continual Anomaly Detection}
With the rapid progress of embodied intelligence and intelligent manufacturing, learning from non-stationary data streams has attracted increasing attention. In industrial visual inspection, continual anomaly detection (CAD) has emerged as an important application of continual learning.

Most existing CAD methods focus on mitigating catastrophic forgetting while introducing new inspection objects \cite{ref17}. For example, UCAD\cite{ref18} enhances feature discriminability via contrastive learning and incorporates prompt mechanisms together with knowledge memory structures to alleviate performance degradation. IUF \cite{ref19} improves the adaptability to new objects by employing object-aware attention and semantic compression loss. More recently, diffusion-based CAD methods have also attracted growing interest. 
CDAD \cite{ref20} reduces gradient interference between old and new tasks via gradient projection, but it still relies on explicit task delineation during updates.

Overall, existing CAD approaches often depend on task identifiers, stage boundaries, or external memory. These assumptions are misaligned with industrial settings, where task transitions are ambiguous, data arrive continuously, and historical samples are difficult to revisit.

\subsection{Task-Free Continual Learning}
Task-free continual learning (TFCL) \cite{ref13} closely matches real-world streaming scenarios, where data arrive continuously and neither training nor inference has access to task boundaries or identifiers. Models must therefore update online under unknown distribution shifts.

To address TFCL, one line of research dynamically models distributional changes and adaptively extends the model structure \cite{ref21}. For instance, Dynamic Cluster Memory (DCM) \cite{ref22} uses knowledge-difference-driven clustering to identify shifts and dynamically grows memory under unsupervised learning. With the advancement of large pretrained models, another line of work freezes the backbone network and introduces lightweight tuning modules to adapt to TFCL scenarios \cite{ref23}. MVP\cite{ref24} leverages the general representation capability of pretrained vision models and proposes instance-level logit masking and contrastive visual prompting, achieving notable performance gains in task-free settings. Similarly, Online-LoRA \cite{ref25} integrates low-rank adaptation modules into Vision Transformers. It automatically detects distribution shifts based on loss-landscape variations, enabling online and task-free parameter-efficient continual learning.

Despite these advances, TFCL remains underexplored for anomaly detection. The lack of anomaly supervision and the need for stable, well-separated feature spaces make existing TFCL methods difficult to transfer directly to this domain.

\subsection{Neural Collapse}
Neural collapse (NC) is an important theoretical finding in deep learning, revealing that deep classification networks tend to form a highly symmetric geometric structure in their feature space during the terminal phase of training. Extensive theoretical studies \cite{ref26}, \cite{ref27} demonstrate that, under simplified models or appropriate regularization, NC corresponds to the global optimum of cross-entropy \cite{ref28} or mean-squared-error \cite{ref29} objectives.

Beyond theoretical analysis \cite{ref30}, \cite{ref31}, researchers have begun to investigate how NC structures may be actively induced in different learning scenarios. In2NeCT \cite{ref32} addresses class-imbalance issues by introducing intra-class and inter-class collapse calibration strategies. These strategies jointly promote feature compactness and separability, mitigating geometric distortion caused by imbalanced data distribution. In recent years, NC has also been explored in continual and online learning settings. FCA \cite{ref26} alleviates catastrophic forgetting in few-shot incremental learning by aligning features with fixed ETF-based class prototypes. Similarly, DYSON \cite{ref33} estimates and updates the optimal geometric structure in an online manner. It guides feature representations toward an ideal collapsed state under task-free incremental learning and achieves competitive performance on natural image classification benchmarks.

These studies collectively indicate that neural collapse is not merely a passive phenomenon emerging during training. Instead, it can serve as an active geometric inductive bias that guides representation learning under non-stationary environments.

\section{Methods}
\subsection{Problem Formulation and Preliminary}
\subsubsection{Task-Free Continual Anomaly Detection}

This work investigates the problem of task-free continual anomaly detection (TF-CAD), \revorange{where training follows a single-pass online continual learning protocol without task boundaries.} Given a chronologically ordered stream
$\mathcal{D}=\{\mathcal{B}_1,\mathcal{B}_2,\ldots,\mathcal{B}_n\}$,
each incoming batch $\mathcal{B}_n=\{x_i\}_{i=1}^{N_n}$ contains only normal training samples. 
\revorange{No real anomalous image or anomaly annotation is available during training. For optimization, synthetic anomalous counterparts are generated from the current normal samples to form an augmented batch, where $y_i=0$ denotes an original normal sample and $y_i=1$ denotes its synthetically perturbed counterpart. Hence, $y_i$ is a synthetic training label rather than a ground-truth anomaly label.}
The data are sequentially fed into the model in a batch-wise manner, while no task boundary or task identifier is provided, and the model is unaware of when or how the underlying distribution changes. Parameter updates use only the current normal batch and its synthetically generated counterparts, without access to historical samples. After training, a unified evaluation is performed over all categories that appear in the data stream. The evaluation assesses whether the learned feature space remains stable while preserving adaptability to continuously evolving distributions. 
\revorange{Here, ``task-free'' refers to the information available to the learner. Although category information is used offline to construct the non-stationary stream, no category label, task identifier, or transition boundary is provided during training. The model therefore performs continual updates without explicit task information.}

\begin{figure*}[!t]
\centering
\includegraphics[width=0.85\textwidth]{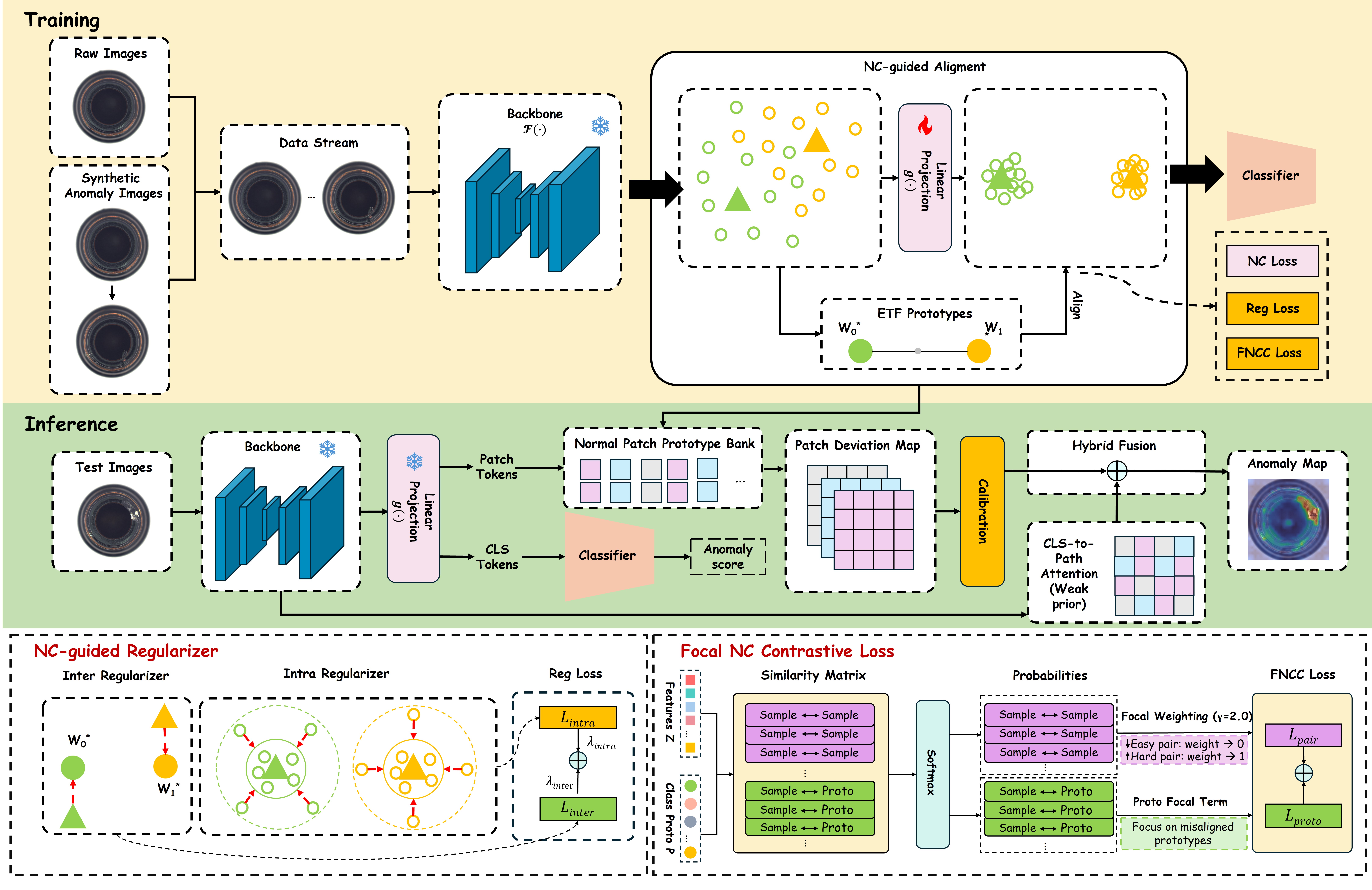}
\caption{The overview of NC-TFAD. \revorange{During training, normal images and their synthetically perturbed counterparts form a task-free data stream} that is fed into a frozen backbone and a \rev{learnable linear projection layer. Specifically, the LayerNorm-normalized \texttt{[CLS]} token from the last Transformer block is projected and $\ell_2$-normalized to form the embedding aligned with fixed ETF prototypes} under NC alignment loss (NC Loss), NC-guided regularization (Reg Loss), and Focal Neural Collapse Contrastive loss (FNCC Loss). \rev{During inference, both the backbone and the trained linear projection layer are frozen, and the same \texttt{[CLS]}-based pathway is used with the fixed ETF prototype classifier to obtain the image-level anomaly score.} \revorange{In parallel, the localization branch constructs a normal patch prototype bank from normal training samples, computes calibrated patch-to-prototype deviation maps for test images, and fuses them with pre-projection CLS-to-patch self-attention as a weak spatial prior to generate the final anomaly map.}}
\label{fig1}
\end{figure*}

\subsubsection{Neural Collapse and Equiangular Tight Frame}
Neural Collapse (NC) refers to a highly symmetric geometric phenomenon that emerges in deep classification networks during the terminal phase of training. When the training error approaches zero and optimization proceeds further, the features of the final layer, class means, and classifier weights spontaneously converge to an analytically characterizable structured configuration. This configuration can typically be represented by a simplex Equiangular Tight Frame (ETF).

Consider a $K$-class classification problem with $K \geq 2$. Let $u_k$ denote the mean feature of class $k$, and assume it is centered such that $\sum_{k=1}^K u_k = 0$. The corresponding simplex ETF can be constructed as

\begin{equation}
    M = \sqrt{\frac{K}{K-1}} \,
    U \left(
    I_K - \frac{1}{K}\mathbf{1}_K\mathbf{1}_K^\top
    \right),
    \label{eq:simplex_etf}
\end{equation}
where
$M=[\mathbf{m}^1,\ldots,\mathbf{m}^K]\in\mathbb{R}^{d\times K}$
is the matrix of class prototypes,
$U\in\mathbb{R}^{d\times K}$ has orthonormal columns satisfying
$U^\top U=I_K$, and
$\mathbf{1}_K\in\mathbb{R}^{K}$ denotes the $K$-dimensional all-ones vector. This structure ensures that all prototypes have equal norm and that the cosine of the angle between any pair of classes equals $-\frac{1}{K-1}$. As a result, the prototypes achieve maximal angular separation and minimal coherence on the unit hypersphere.

The Neural Collapse phenomenon can be summarized as:
\begin{itemize}
    \item \textbf{NC1:} Features of samples from the same class converge to their class mean;
    \item \textbf{NC2:} Class means converge to the vertices of a simplex ETF, with equal norm and equal inter-class angles;
    \item \textbf{NC3:} Classifier weight vectors align directionally with their corresponding class means;
    \item \textbf{NC4:} The classifier decision rule becomes equivalent to a nearest-class-mean decision in Euclidean space.
\end{itemize}

\revorange{
In the binary normal--anomaly setting considered in this work, the simplex ETF yields two antipodal prototypes that define a fixed and maximally separated geometric reference for the two anomaly states. This property is well suited to task-free continual anomaly detection, where non-stationary data streams can progressively distort the representation space and shift the decision boundary. Motivated by this observation, we adopt the NC geometry as a category-agnostic structural prior to maintain a consistent normal--anomaly organization throughout continual adaptation. Specifically, the fixed ETF prototypes provide stable reference directions, while NC-inspired class-mean alignment and intra-class compactness further constrain the evolving embeddings to remain discriminative under distribution shifts.
}

\subsection{Framework Overview}
Fig.\ref{fig1} illustrates the overall architecture of NC-TFAD. The framework consists of a training process and an inference-time localization process.

During training, each batch contains both normal and synthetic anomalous samples with labels $y_i \in \{0,1\}$. An input image $x_i$ is first processed by the frozen backbone $\mathcal{F}(\cdot)$ to obtain the final-layer token sequence. For the image-level training branch, we use the LayerNorm-normalized \texttt{[CLS]} token as the global image representation, rather than pooling all patch tokens:

\begin{equation}
    f_i^{\mathrm{cls}} = \operatorname{LN}\!\left(\mathcal{F}_{\mathrm{cls}}(x_i)\right),
    \label{eq:feature_extraction}
\end{equation}
where $\mathcal{F}_{\mathrm{cls}}(x_i)$ denotes the \texttt{[CLS]} token extracted from the last Transformer block. This \texttt{[CLS]} feature is then projected into a learnable embedding space through a linear projection layer $g(\cdot)$:

\begin{equation}
   \revorange{
    z_i =
    \frac{g(f_i^{\mathrm{cls}})}
    {\|g(f_i^{\mathrm{cls}})\|_2},
    \quad z_i \in \mathbb{R}^d,
    \quad \|z_i\|_2=1.
    }
    \label{eq:embedding_projection}
\end{equation}

Here, $d$ denotes the embedding dimension and the resulting embeddings form the set $Z = \{z_1, z_2, \dots, z_n\}$. \revorange{The ETF prototypes are also unit-normalized, i.e., $\|p_k\|_2=1$. Hence, all subsequent sample--sample and sample--prototype dot products are evaluated in the normalized embedding space and correspond to cosine similarities.} This embedding space serves as a unified domain for geometric alignment, regularization, and contrastive learning.

Within this space, two fixed ETF prototypes $\mathcal{P} = \{p_0, p_1\}$ are pre-constructed according to Eq.~\eqref{eq:simplex_etf}, corresponding to the normal and anomalous classes. \rev{Here, the prototype index represents the anomaly state rather than the industrial object category. Specifically, $p_0$ and $p_1$ are shared by normal and anomalous samples, respectively, across all object categories. Category identities are not provided to the model and are not used to construct category-specific prototypes. This category-agnostic design provides a fixed normal--anomalous geometric reference that does not require task or category identification as the data distribution evolves.} \revorange{These prototypes remain fixed during training and provide stable geometric references throughout the stream. The learnable projection layer is optimized with three complementary objectives: NC Loss aligns individual embeddings with their ETF prototypes; Reg Loss preserves the global normal--anomaly geometry through class-mean alignment and intra-class compactness; and FNCC Loss improves sample-level discriminability by modeling sample--sample and sample--prototype relations with emphasis on hard pairs. Together, they maintain a structured and discriminative embedding space under non-stationary distributions. The fixed ETF prototypes also serve as geometric classifier weights, yielding predictions via}:

\begin{equation}
    \hat{y}_i = \arg\max_{k \in \{0,1\}} z_i^\top p_k.
    \label{eq:prediction}
\end{equation}

\rev{
During inference, both the pretrained backbone and the learned linear projection layer are frozen. Given a test image $x$, the representation used for image-level anomaly detection is obtained by extracting the LayerNorm-normalized \texttt{[CLS]} token from the last Transformer block, followed by the same $384$-to-$384$ linear projection and $\ell_2$ normalization used during training. The resulting embedding is then compared with the fixed ETF prototypes to obtain the image-level anomaly prediction and anomaly confidence.}
\revorange{
For pixel-level localization, NC-TFAD uses the patch tokens extracted by the frozen backbone to construct a normal patch prototype bank from normal training samples. At inference time, each test patch is scored by its calibrated deviation from the nearest normal prototype, yielding a prototype-based anomaly response. This response is further fused with the original backbone CLS-to-patch self-attention as a weak spatial prior. We use the pre-projection \texttt{[CLS]} attention because it directly reflects token-level interactions in the frozen vision backbone, whereas the projected \texttt{[CLS]} embedding is optimized for image-level ETF discrimination and may discard spatial correspondence useful for localization.
}

\subsection{Synthetic Anomaly Generation}
\revred{In industrial anomaly detection, training data typically contain only normal samples, whereas the proposed NC-based normal--anomaly geometry requires reference directions for both states. To instantiate the otherwise unavailable anomalous direction, we generate synthetic anomalies from normal images and use them exclusively as auxiliary geometric anchors during training. These synthetic samples provide weak image-level abnormality cues that enable the projection layer to establish a discriminative normal--anomaly geometry in the absence of real anomaly supervision.}

\begin{figure}[!t]
\centering
\includegraphics[width=0.5\textwidth]{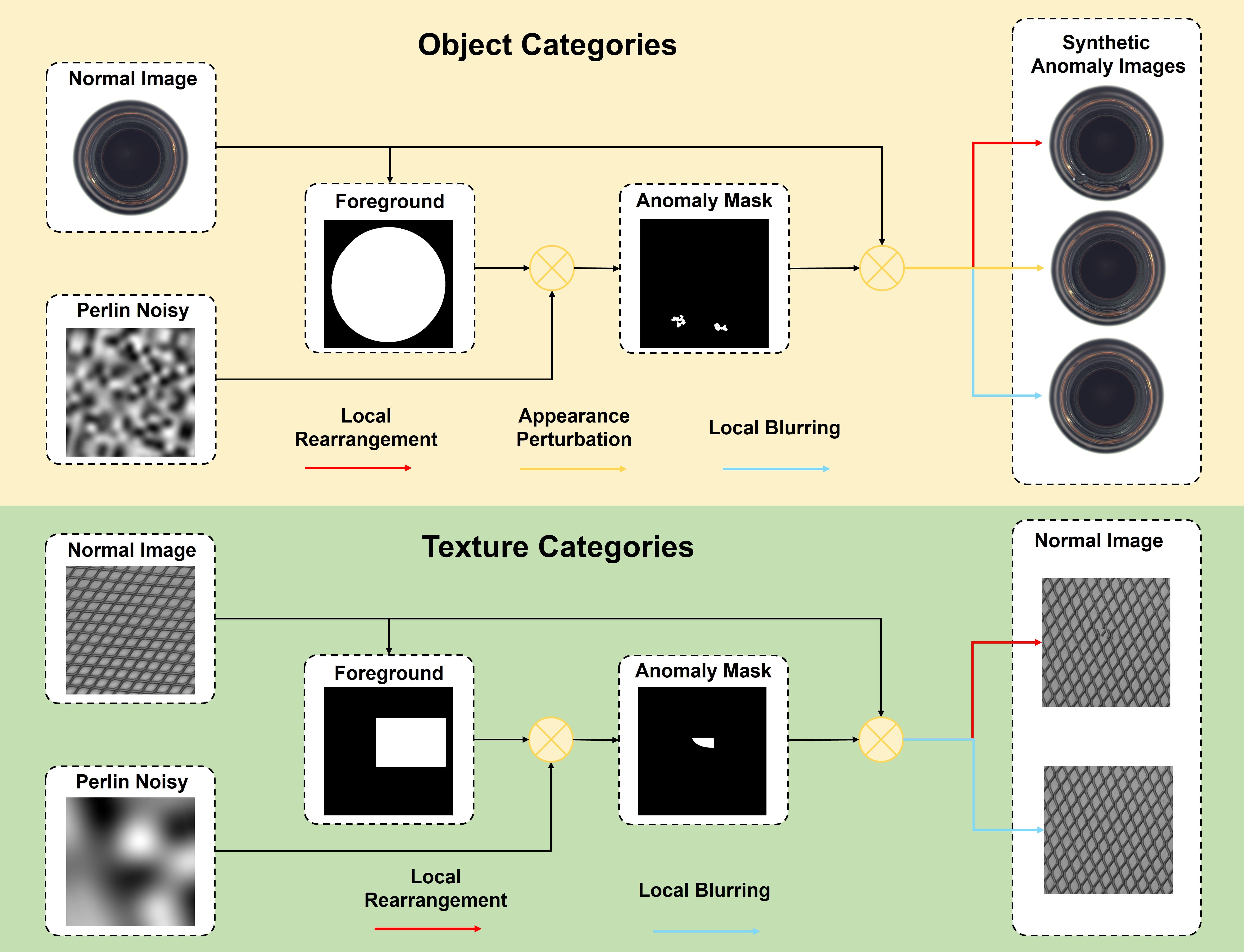}
\caption{\revorange{Pipeline for synthetic anomaly generation.}}
\label{fig2}
\end{figure}

As shown in Fig.\ref{fig2}, following prior work \cite{ref6}, \cite{ref35}, \cite{ref36}, the data are divided into object categories and texture categories. For object categories, a coarse foreground is obtained by grayscale thresholding (Otsu or triangle, with inversion when necessary) and refined via morphological operations. For texture data, the entire image is treated as foreground, as defects may appear anywhere on the surface. 

To determine candidate anomaly regions, we employ Perlin noise to generate irregular spatial patterns with natural and diverse shapes. These candidate anomaly regions are intersected with the foreground to ensure that anomalies are injected only within semantically valid areas, while regions outside the foreground are discarded. The resulting intersection defines the final anomaly mask. Within this mask, we then apply one of three local perturbations with equal probability: appearance perturbation that alters color and intensity statistics, local blurring that disrupts fine-grained texture patterns, or local rearrangement that breaks spatial continuity by shuffling subregions. For texture categories, we restrict the operation set to local blurring and local rearrangement. For object categories, we further impose constraints on the number and size of connected anomalous regions to avoid overly fragmented artifacts, and apply smooth alpha blending along mask boundaries to reduce visual discontinuities.

\rev{
The resulting perturbed images serve as synthetic anomalous samples for image-level geometric learning. The spatial mask is used only to constrain local perturbations and is discarded after synthesis, without being provided to the backbone or used as pixel-level supervision. Since the ViT \texttt{[CLS]} token aggregates information from patch representations through self-attention, localized perturbations can still influence the global representation, enabling the projection layer to learn image-level normal--anomaly separation without explicit anomaly-location supervision.
}

\subsection{NC-Guided Regularization}

In task-free continual learning, models must adapt to non-stationary streams without degrading previously learned decision structures. We therefore propose an NC-guided geometric regularization that enforces inter-class alignment and intra-class compactness in the embedding space. This regularization stabilizes the embedding geometry and mitigates catastrophic forgetting and representation drift.

\subsubsection*{Inter-Class Alignment Regularization}
NC theory indicates that, in the terminal phase of training, class means converge to a simplex ETF. Inspired by this property, we introduce a set of fixed directional priors in the embedding space:

\begin{equation}\mathbf{W}^*=
\begin{bmatrix}
\mathbf{w}_0^{*\top} \\
\mathbf{w}_1^{*\top}
\end{bmatrix}\in\mathbb{R}^{2\times d},\quad\|\mathbf{w}_k^*\|_2=1.\end{equation}
where $\mathbf{w}_k^*$ denotes the target alignment direction of class $k \in \{0,1\}$, obtained from the ETF prototype 
$\mathcal{P} = \{p_0, p_1\}$ in Eq.~\eqref{eq:simplex_etf}. Let $S \subseteq \{0,1\}$ denote the set of classes appearing in the current batch. For any $k \in S$, the batch mean embedding is computed as

\begin{equation}
    \mathbf{u}_k = \frac{1}{|\mathcal{B}_k|} \sum_{i \in \mathcal{B}_k} \mathbf{z}_i,
    \label{eq:batch_mean}
\end{equation}
where $\mathcal{B}_k$ is the index set of samples belonging to class $k$. The similarity between the class mean and directional prior is defined as

\begin{equation}
    u_{k \to l} = \mathbf{u}_k^\top \mathbf{w}_l^*, \quad k, l \in S.
    \label{eq:similarity}
\end{equation}

Based on the similarity matrix, the inter-class alignment loss is formulated as a row-wise softmax cross-entropy:

\begin{equation}
    L_{\text{inter}} = \frac{1}{|S|} \sum_{k \in S} \left( -\log \frac{\exp(u_{k \to k})}{\sum_{l \in S} \exp(u_{k \to l})} \right).
    \label{eq:inter_loss}
\end{equation}

This loss encourages each class mean to align with its corresponding prior direction. When $|S| < 2$ or the prior is unavailable, this term is omitted to ensure optimization stability.

\subsubsection*{Intra-Class Compactness Regularization}
Beyond inter-class alignment, NC also reveals a trend of within-class feature aggregation and variance collapse. To enhance intra-class consistency and improve robustness to distribution shift and noise, we introduce an intra-class compactness constraint.

The pairwise similarity between two distinct samples within the same class is defined as

\begin{equation}
    d_{ij}^{(k)} = z_i^\top z_j, \quad i \ne j,\, i,j \in \mathcal{B}_k.
    \label{eq:pairwise_sim}
\end{equation}
For each sample $z_i$, similarities to all other samples from the same class are aggregated as
\begin{equation}
    a_i = -\log\left(\frac{1}{n_k - 1} \sum_{j \neq i} \exp{d_{ij}^{(k)}}\right).
    \label{eq:aggregated_sim}
\end{equation}
Based on this aggregation, the intra-class compactness loss is given by

\begin{equation}
    L_{\text{intra}} =  \sum_{k \in S} \frac{1}{n_k^\zeta} \frac{1}{n_k} \sum_{i \in \mathcal{B}_k} a_i .
    \label{eq:intra_loss}
\end{equation}
Here, $\zeta = 1$ controls inverse-frequency re-weighting to mitigate within-batch class imbalance and increase sensitivity to minority (anomalous) classes. This constraint increases the overall similarity among samples from the same class, encouraging tighter intra-class clusters and suppressing representation scattering caused by drift and noise.

\subsubsection*{Regularization Objective}

The overall NC-guided regularization objective is defined as

\begin{equation}
    L_{\text{reg}} = \lambda_{\text{inter}} L_{\text{inter}} + \lambda_{\text{intra}} L_{\text{intra}},
    \label{eq:regularization_objective}
\end{equation}
where $\lambda_{\text{inter}}$ and $\lambda_{\text{intra}}$ balance the contributions of the two terms. 
\revorange{The ETF directions are used as a fixed geometric inductive bias for organizing the streaming embeddings, and the formulation does not require the embeddings to attain an exact neural-collapse configuration.}

\subsection{Focal NC Contrastive Loss}

In task-free continual learning, some samples may still remain close to the decision boundary or deviate substantially from their target prototypes. To address this issue, we introduce a Focal Neural Collapse Contrastive loss (FNCC Loss), which enhances sample-level discriminability via contrastive learning while emphasizing hard-to-align samples within the NC-induced geometry.

\subsubsection*{Similarity Measurement and Unified Partition Function}

We first define similarity metrics for both sample--sample and sample--prototype relations:

\begin{align}
    s_{ij} &= \frac{\mathbf{z}_i^\top \mathbf{z}_j}{\tau}, \quad i,j \in \mathcal{B},\, i \ne j, \label{eq:sim_sample_sample} \\
    t_{ik} &= \frac{\mathbf{z}_i^\top {p}_k}{\tau}, \quad i \in \mathcal{B},\, k \in S, \label{eq:sim_sample_prototype}
\end{align}
where $\tau > 0$ is the temperature parameter and $p_k$ denotes the prototype of class $k$. To unify both relations within a single normalized probability space, the log partition function for sample $i$ is given by

\begin{equation}
    G_i =
    \begin{cases}
        \displaystyle  \sum_{j=1}^{n} \exp{s_{ij}} + \sum_{k=1}^{|\mathcal{P}|} \exp{t_{ik}}, & \mathcal{P} \ne \emptyset, \\
        \displaystyle \sum_{j=1}^{n} \exp{s_{ij}}, & \text{otherwise}.
    \end{cases}
    \label{eq:log_partition}
\end{equation}
Here, $\mathcal{P}$ represents the set of class prototypes involved in the current batch; when no prototype is available, 
$\mathcal{P} = \emptyset$. The conditional probability of sample $j$ given anchor sample $i$ is then defined as

\begin{equation}
    P(j \mid i) = \exp\left(s_{ij} - \log G_i\right).
    \label{eq:conditional_prob}
\end{equation}

\subsubsection*{Positive-Pair Term}

Positive pairs are defined as samples belonging to the same class but not identical: $\Omega^+ = \{(i,j) \mid y_i = y_j, i \ne j\}$. To focus optimization on hard positive pairs, we introduce a focal modulation. The sample-pair focal contrastive loss is formulated as

\begin{equation}
    L_{\text{pair}} = -\frac{1}{|\Omega^+|} \sum_{(i,j) \in \Omega^+} \left(1 - P(j \mid i)\right)^\gamma \log\left(\max(P(j \mid i), \varepsilon)\right),
    \label{eq:focal_pair_loss}
\end{equation}
where $\gamma \geq 0$ is the focusing parameter and $\varepsilon = 10^{-8}$ ensures numerical stability. If $|\Omega^+| = 0$, we set $L_{\text{pair}} = 0$. This term assigns higher weights to low-confidence positive pairs, adaptively emphasizing samples that lie near decision boundaries or are affected by distribution shift.

\subsubsection*{Prototype-Assignment Term}

Let $\pi(\cdot)$ denote the mapping from class labels to prototypes. For a sample $i$ with $\pi(y_i) = p_k$, the probability of being assigned to its correct prototype is

\begin{equation}
    R(i) = \exp\left(t_{i,\pi(y_i)} - \log G_i\right).
    \label{eq:prototype_assignment_prob}
\end{equation}

Based on this probability, the prototype-guided focal term is given by

\begin{equation}
    L_{\text{proto}} = -\frac{1}{|\mathcal{H}|} \sum_{i \in \mathcal{H}} \left(1 - R(i)\right)^\gamma \log R(i),
    \label{eq:focal_proto_loss}
\end{equation}
where $\mathcal{H} = \{i \in \mathcal{B} \mid \pi(y_i) \ne \emptyset\}$. If $\mathcal{H}$ is empty, we set $L_{\text{proto}} = 0$. This term strengthens sample--prototype alignment and encourages clearer prototype-centered cluster structures.

\subsubsection*{Final Contrastive Objective}
The proposed Focal NC Contrastive Loss is defined as

\begin{equation}
    L_{\text{FNCC}} = L_{\text{pair}} + L_{\text{proto}}.
    \label{eq:fncc_loss}
\end{equation}

By re-weighting probabilities via the focal factor $\gamma$, FNCC adaptively emphasizes hard positive pairs and ambiguous prototype assignments. This design enhances the discriminability and robustness of the embedding space under continual distribution shifts.

\subsection{Anomaly Localization}

\subsubsection*{\revorange{Normal Patch Prototype Bank}}
\revorange{The localization branch is used only for pixel-level heatmap generation and does not change the image-level training or evaluation pathway. To obtain a more precise localization signal than attention alone, we construct a compact normal patch prototype bank using only normal samples from the training stream. Specifically, for a normal training image $x$, the frozen backbone extracts patch tokens $\{\mathbf{r}_i\}_{i=1}^{N}$ from the last Transformer block, where $N$ is the number of spatial patches. The patch tokens are $\ell_2$-normalized and summarized into a compact set of normal patch prototypes:
\begin{equation}
    \mathcal{C}
    =
    \{\mathbf{c}_m\}_{m=1}^{N_c},
    \qquad
    \|\mathbf{c}_m\|_2=1,
    \label{eq:patch_proto_bank}
\end{equation}
where $N_c$ denotes the number of normal patch prototypes. 
The prototype bank $\mathcal{C}$ is constructed by online sampling followed by lightweight $k$-means clustering over normal patch tokens. 
Only normal training patches are used in this process; synthetic anomalies, test images, and pixel-level masks are excluded from prototype construction. 
This design preserves the normal-only supervision setting of industrial anomaly detection while remaining compatible with the task-free continual learning protocol.}

\subsubsection*{\revorange{Prototype-Calibrated Patch Deviation Map}}
\revorange{For a test image, each normalized patch token $\mathbf{r}_i$ is compared with the normal patch prototype bank. The patch-level deviation score is defined as the cosine distance to the nearest normal prototype:}
\begin{equation}
    d_i
    =
    1-
    \max_{1\leq m\leq N_c}
    \mathbf{r}_i^\top \mathbf{c}_m .
    \label{eq:patch_proto_distance}
\end{equation}
\revorange{Rather than applying per-image min--max normalization, which may amplify small score variations and yield spuriously high responses even for normal images, we calibrate patch-level deviations using statistics estimated from normal training data. Specifically, after constructing the prototype bank $\mathcal{C}$, we compute the nearest-prototype cosine distance for every normal training patch using Eq.~\eqref{eq:patch_proto_distance}. Let $q_{95}$ denote the 95th percentile of these normal nearest-prototype distances over the training stream. A test-patch distance $d_i$ is then calibrated relative to this normal reference distribution using a localization temperature $\tau_l$:}
\begin{equation}
    \mathbf{S}_i^{\mathrm{proto}}
    =
    \sigma\!\left(\frac{d_i-q_{95}}{\tau_l}\right).
    \label{eq:patch_proto_calibration}
\end{equation}
\revorange{The patch responses are rearranged into a spatial grid and bilinearly upsampled to the input resolution, yielding $\mathbf{S}^{\mathrm{proto}}\in[0,1]^{H\times W}$. This training-distribution calibration improves pixel-level ranking stability under severe foreground--background imbalance.}

\subsubsection*{\revorange{Attention-Prior Hybrid Fusion}}
\revorange{Although prototype deviations provide discriminative local anomaly evidence, they may remain sensitive to benign appearance variations. We therefore retain backbone self-attention as a weak spatial prior. CLS-to-patch attention is extracted from the frozen backbone, fused across attention heads, rearranged into a spatial map, and upsampled to obtain $\mathbf{A}\in[0,1]^{H\times W}$. The resulting high-confidence anomaly map is defined as}
\begin{equation}
    \mathbf{H}^{\mathrm{prec}}
    =
    \mathbf{S}^{\mathrm{proto}}
    \odot
    (\epsilon+\mathbf{A})^{\gamma_l},
    \label{eq:hybrid_precision_map}
\end{equation}
\revorange{where $\epsilon$ prevents excessive suppression and $\gamma_l$ controls the strength of the attention prior. To extend spatial support while limiting the influence of smoothing on high-confidence anomaly responses, we further construct a low-confidence support map}
\begin{equation}
    \mathbf{H}^{\mathrm{cov}}
    =
    \mathrm{AvgPool}(\mathbf{H}^{\mathrm{prec}})
    \odot \mathbf{A},
    \label{eq:hybrid_coverage_map}
\end{equation}
\revorange{and obtain the final localization map as}
\begin{equation}
    \mathbf{H}
    =
    \max\!\left(\mathbf{H}^{\mathrm{prec}}, \rho\,\mathbf{H}^{\mathrm{cov}}\right),
    \label{eq:hybrid_final_map}
\end{equation}
\revorange{where $\rho\in(0,1)$ assigns lower confidence to the expanded support regions. This dual-map fusion preserves high-confidence anomaly responses while extending spatial support for region-level localization.}

\subsubsection*{\revorange{Connected-Region Filtering}}
\revorange{Finally, isolated anomaly responses are further suppressed using Top-$k$ connected-region filtering. For each image, high- and low-response thresholds are defined as}
\begin{equation}
    \theta_h = \rho_h \max(\mathbf{H}), \quad \theta_l = \rho_l \max(\mathbf{H}), \quad 0 < \rho_l < \rho_h < 1.
    \label{eq:thresholds}
\end{equation}
\revorange{Connected components above $\theta_h$ are used as confident seeds, and low-threshold components are retained only when they overlap with these seeds. Responses outside the retained support are attenuated by $\eta$. This post-processing operates exclusively on the pixel-level heatmap and does not affect image-level prediction.}

\section{Experiments}

\subsection{Datasets}
\textit{MVTec AD}: MVTec AD\cite{ref37} is one of the most widely used benchmarks for industrial anomaly detection. It contains 15 categories in total, including 10 object categories and 5 texture categories. The dataset consists of 4,096 normal images and 1,258 anomalous images. These anomalous samples cover 73 types of realistic defects, such as scratches, dents, structural deformations, and surface damages.

\textit{VisA}: VisA\cite{ref38} comprises 12 subsets corresponding to 12 industrial object categories. In total, it contains 10,821 images, including 9,621 normal samples and 1,200 anomalous samples. The anomalies span a broad range of defect types, including scratches, dents, stains, cracks, and other surface defects, as well as structural anomalies such as misalignment and missing components.

\subsection{Evaluation Metrics}
To evaluate anomaly detection under the task-free continual learning setting, we follow a unified evaluation protocol at the end of the data stream \cite{ref7}, \cite{ref18}. Specifically, Image-level Area Under the Receiver Operating Characteristic curve (I-AUROC) is used to assess the ability to distinguish between normal and anomalous images, while Pixel-level AUROC (P-AUROC) is used to evaluate anomaly localization performance. Given the severe class imbalance in industrial scenarios, AUROC alone can be insufficient, so we additionally report image-level Average Precision (I-AP) and pixel-level Average Precision (P-AP) to more accurately assess detection and localization under highly imbalanced conditions. \revorange{For pixel-level localization, we further report the Area Under the Per-Region Overlap curve (P-AUPRO), which measures the overlap between predicted anomaly regions and individual ground-truth defect regions across different thresholds, providing a complementary assessment of region-level localization quality.}

\subsection{Implementation Details}
We adopt a \revred{DINO} ViT-S/8 pretrained on ImageNet~\cite{ref39,ref40} as the frozen backbone and resize all images to $224 \times 224$. \rev{For image-level anomaly detection, only the final LayerNorm output of the \texttt{[CLS]} token is used as the backbone representation. The 384-dimensional feature is passed through a bias-free linear projection layer ($384\rightarrow384$) without activation, BatchNorm, or Dropout, introducing 147,456 trainable parameters. The projected feature is $\ell_2$-normalized before NC-guided optimization and classification with fixed ETF prototypes.} Training follows the Online Task-Free Continual Learning (OTFCL) protocol, where each sample is accessed once. We use a batch size of 64 and Adam optimizer with an initial learning rate of $2\times10^{-5}$ and weight decay of $5\times10^{-6}$. The loss weights are set to $\gamma=2.0$, $\lambda_{\text{inter}}=0.9$, and $\lambda_{\text{intra}}=0.5$.

\revorange{The localization branch is independent of the image-level linear projection layer and introduces no trainable parameters. Normalized patch tokens from the frozen backbone are summarized into a normal patch prototype bank with $N_c=64$ prototypes using 5,000 normal patch tokens. Pixel-level scores are computed by nearest-prototype cosine distance and calibrated with the 95-th percentile normal distance ($\tau_l=0.02$). The attention-prior fusion uses $\epsilon=0.2$, $\gamma_l=0.5$, $\rho=0.35$, and a $5\times5$ average-pooling window. Connected-region filtering uses $\rho_h=0.7$, $\rho_l=0.2$, $k=4$, and $\eta=0.2$. The localization branch only affects pixel-level heatmaps, while image-level scores are obtained exclusively from the \texttt{[CLS]}-based ETF classifier.} 

\revorange{For stream construction, samples are shuffled within each category and arranged into category-contiguous segments according to a fixed random category order. Consecutive segments are directly concatenated to create abrupt distribution shifts, while category labels remain unavailable to the model during training. Specifically, the predefined category order for MVTec AD is Tile, Transistor, Toothbrush, Leather, Carpet, Wood, Pill, Hazelnut, Capsule, Cable, Bottle, Zipper, Grid, Screw, and Metal Nut. For VisA, the order is Chewing Gum, Macaroni1, Macaroni2, PCB4, Cashew, Capsules, PCB2, Candle, Pipe Fryum, Fryum, PCB3, and PCB1. All methods are evaluated using the same predefined category orders.} Experiments are implemented in PyTorch on a single NVIDIA RTX 4090 GPU.

\subsection{Baselines}
\revred{We compare NC-TFAD with representative methods from both continual learning and industrial anomaly detection to provide a comprehensive evaluation under the considered task-free continual anomaly detection setting. The continual learning baselines include ER~\cite{ref41}, DualPrompt~\cite{ref42}, L2P~\cite{ref43}, MVP~\cite{ref24}, FCA~\cite{ref26}, DCM~\cite{ref22}, DYSON~\cite{ref33}, and Online-LoRA~\cite{ref25}, which represent a range of replay-based, prompt-based, geometry-guided, dynamic-memory, and parameter-efficient continual learning strategies. The industrial anomaly detection baselines include UniAD~\cite{ref7}, UCAD~\cite{ref18}, IUF~\cite{ref19}, CDAD~\cite{ref20}, and IB-IUMAD~\cite{ref_ibiumad}, covering representative unified and continual anomaly detection frameworks. Together, these methods enable evaluation against both general continual learning strategies and approaches specifically designed for industrial anomaly detection.}

\revred{Recent vision--language-based anomaly detection methods, such as WinCLIP~\cite{ref_winclip} and its variants, primarily operate under zero-shot or few-shot protocols, where the target domain requires no continual training or only a small number of reference samples. This setting differs from OTFCL, in which normal samples from evolving target-domain distributions are processed sequentially for single-pass continual updates without task boundaries or category identifiers. Owing to this protocol difference, vision--language-based methods are considered related approaches rather than direct quantitative baselines under OTFCL.} 
For ER, the replay memory is fixed at 500 samples. 
\revorange{All applicable methods are evaluated under the same OTFCL stream and evaluation protocol.}

\subsection{Experimental Results}

\subsubsection*{Detection Results on the MVTec AD Dataset}
\revred{
As shown in Table~\ref{tab:table1}, NC-TFAD achieves the best average image-level detection performance on MVTec AD and maintains strong results across diverse object and texture categories. Its advantage is consistently observed over both general continual learning methods and approaches specifically developed for industrial anomaly detection, indicating that directly adapting either paradigm is insufficient to handle the normal-only, task-free, and non-stationary setting considered in this work. }

\revred{
DYSON is an NC-based method originally developed for online multiclass classification, where observed semantic labels directly define class-dependent geometric structures. Under the task-free anomaly detection setting considered here, the training stream contains only normal samples and therefore provides no observed anomalous counterpart for constructing a normal--anomaly geometry. Introducing synthetic anomalies enables DYSON* to operate under a two-class training formulation, but its original optimization remains designed for online multiclass classification. In contrast, NC-TFAD constructs a category-agnostic normal--anomaly ETF reference and jointly incorporates class-mean alignment, intra-class compactness, and FNCC-based sample--sample and sample--prototype optimization. As shown in Table~\ref{tab:table1}, NC-TFAD achieves substantially higher average detection performance than DYSON*, indicating the effectiveness of adapting NC-based geometry and continual optimization specifically to industrial anomaly detection under non-stationary streams.
}

\begin{table*}[t]
\centering
\caption{\revred{Image-level anomaly detection performance (\%, I-AUROC / I-AP) on MVTec AD under the OTFCL setting.}}
\begin{adjustbox}{width=\textwidth, center}
\label{tab:table1}
\begin{tabular}{l|ccccccccccccccc|c}
\toprule
Method & Bottle & Cable & Capsule & Carpet & Grid & Hazelnut & Leather & Metal\_nut & Pill & Screw & Tile & Toothbrush & Transistor & Wood & Zipper & Average \\
\midrule
ER
& 28.7/70.7 & 36.2/58.3 & 70.4/92.1 & 69.4/90.2
& 50.0/76.4 & 29.1/52.2 & 65.0/86.2 & 34.1/73.2
& 45.4/85.3 & 29.7/61.4 & 67.3/86.0 & 65.6/85.5
& 53.3/45.6 & 67.2/86.9 & 42.1/77.9 & 50.2/76.3 \\

DualPrompt
& 79.2/93.6 & 73.2/78.9 & \textbf{75.8}/\textbf{94.0} & 67.4/88.8
& 62.7/79.6 & 43.9/63.3 & 34.9/69.3 & 47.2/79.9
& 59.0/89.3 & 58.3/80.4 & 68.8/86.0 & 47.2/70.3
& 55.0/52.0 & 68.3/90.1 & 66.5/88.6 & 60.5/80.3 \\

UniAD
& 53.9/80.8 & 41.9/58.0 & 50.6/84.5 & 92.7/97.0
& 76.4/89.4 & 83.3/89.5 & 86.6/95.2 & 42.5/82.0
& 53.4/86.1 & 51.9/75.6 & 64.8/84.7 & 56.1/77.0
& 52.4/49.6 & 88.7/96.4 & 50.8/85.0 & 63.1/82.0 \\

L2P
& 89.1/96.8 & 59.5/73.3 & 43.2/82.0 & 56.7/81.8
& 56.7/77.2 & 67.6/83.3 & 38.9/74.9 & 59.7/\textbf{96.6}
& 50.1/87.3 & 48.8/72.4 & 73.2/85.3 & 28.9/60.3
& 55.2/55.1 & 56.5/78.1 & 67.8/88.7 & 56.8/78.9 \\

MVP
& 66.4/87.5 & 63.0/72.3 & 70.9/92.6 & 32.2/71.3
& 57.9/80.1 & 40.5/63.8 & 41.5/76.7 & 61.7/89.4
& 50.6/88.1 & 33.2/64.3 & 37.5/67.9 & 55.6/78.1
& \textbf{67.0}/61.3 & 45.5/78.8 & 56.0/85.7 & 52.0/77.2 \\

FCA*
& 63.6/86.3 & 43.6/43.6 & 65.5/87.9 & 51.9/79.6
& 64.9/81.0 & 30.9/52.6 & 30.0/66.0 & 41.3/77.4
& 66.4/91.0 & 98.6/99.6 & 36.9/67.7 & 48.9/74.9
& 66.7/\textbf{65.4} & 45.7/80.8 & 55.1/83.1 & 54.0/77.0 \\

DCM
& 49.2/77.6 & 49.8/64.3 & 18.6/84.4 & 59.4/61.7
& 83.1/91.7 & 64.5/78.5 & 72.0/89.7 & 34.0/75.3
& 49.9/82.8 & \textbf{99.9}/\textbf{99.9} & 39.7/64.4 & 49.4/79.3
& 34.3/33.3 & 66.8/86.4 & 24.7/68.1 & 53.0/75.8 \\

DYSON
& 50.0/75.9 & 50.0/61.3 & 50.0/82.6 & 50.0/76.1
& 50.0/73.1 & 50.0/63.6 & 50.0/74.2 & 50.0/80.9
& 50.0/84.4 & 50.0/74.4 & 50.0/71.8 & 50.0/71.4
& 50.0/40.0 & 50.0/75.9 & 50.0/78.8 & 50.0/72.3 \\

DYSON*
& 95.6/98.7 & 58.6/75.5 & 38.7/82.4 & 76.8/93.1
& 78.6/91.6 & 74.2/85.3 & \textbf{100.0}/\textbf{100.0} & 55.4/85.1
& 63.6/91.5 & 61.0/78.9 & 80.6/93.6 & 86.7/95.8
& 55.3/48.0 & 70.4/88.6 & 55.3/86.2 & 70.1/86.3 \\

Online-LoRA
& 80.6/94.2 & 65.5/78.4 & 52.0/84.7 & 83.8/95.1
& 54.0/79.5 & 32.4/56.3 & 74.8/90.8 & 55.1/82.7
& 40.9/80.7 & 9.2/55.3 & 77.9/90.6 & 65.6/86.2
& 58.3/53.9 & 87.4/96.0 & \textbf{83.3}/\textbf{94.9} & 61.4/81.3 \\

\revred{IUF}
& \revred{59.60/85.02}
& \revred{55.49/59.97}
& \revred{70.08/76.19}
& \revred{97.07/99.20}
& \revred{70.76/87.57}
& \revred{88.75/94.35}
& \revred{97.89/99.30}
& \revred{64.71/78.57}
& \revred{55.29/86.83}
& \revred{52.61/72.27}
& \revred{84.05/93.71}
& \revred{50.00/77.48}
& \revred{56.46/43.00}
& \revred{\textbf{95.35}/\textbf{98.73}}
& \revred{64.97/89.65}
& \revred{70.87/82.79} \\

\revred{CDAD}
& \revred{57.86/83.25}
& \revred{54.16/67.61}
& \revred{52.29/82.78}
& \revred{61.60/80.28}
& \revred{51.55/77.72}
& \revred{70.75/78.31}
& \revred{76.63/90.24}
& \revred{51.22/82.35}
& \revred{53.93/86.07}
& \revred{62.16/67.52}
& \revred{60.53/80.33}
& \revred{55.28/80.51}
& \revred{51.46/43.22}
& \revred{86.23/94.65}
& \revred{60.27/87.01}
& \revred{60.39/78.79} \\

\revred{UCAD}
& \revred{65.40/88.22}
& \revred{\textbf{78.71}/\textbf{83.86}}
& \revred{52.05/85.13}
& \revred{94.62/98.41}
& \revred{79.28/92.36}
& \revred{91.29/94.21}
& \revred{99.66/99.88}
& \revred{50.10/82.32}
& \revred{53.22/88.99}
& \revred{56.45/80.64}
& \revred{79.69/91.03}
& \revred{73.89/90.57}
& \revred{50.75/49.29}
& \revred{92.28/97.00}
& \revred{63.71/88.75}
& \revred{72.07/87.38} \\

\revred{IB-IUMAD}
& \revred{58.97/84.98}
& \revred{56.54/59.23}
& \revred{68.45/76.84}
& \revred{\textbf{97.27}/\textbf{99.24}}
& \revred{69.09/87.32}
& \revred{87.39/93.67}
& \revred{98.27/99.43}
& \revred{66.62/77.66}
& \revred{55.67/87.28}
& \revred{51.44/73.10}
& \revred{86.83/94.91}
& \revred{50.56/77.94}
& \revred{58.67/42.24}
& \revred{\textbf{95.35}/\textbf{98.73}}
& \revred{66.44/90.10}
& \revred{71.17/82.84} \\

Ours
& \textbf{99.8}/\textbf{99.2}
& 65.8/77.2
& 69.4/92.7
& 88.4/96.9
& \textbf{87.8}/\textbf{95.6}
& \textbf{93.3}/\textbf{96.7}
& \textbf{100.0}/\textbf{100.0}
& \revred{\textbf{78.8}/94.7}
& \textbf{84.6}/\textbf{96.9}
& 68.1/87.7
& \textbf{96.1}/\textbf{98.7}
& \textbf{91.1}/\textbf{96.8}
& 57.9/56.8
& 81.3/94.7
& 69.6/90.9
& \textbf{82.1}/\textbf{91.8} \\
\bottomrule
\end{tabular}
\end{adjustbox}

\smallskip
\raggedright
\footnotesize
$^*$FCA* and DYSON* are NC-based continual learning methods; synthetic anomaly samples were used to meet their two-class training requirement.
\end{table*}

\subsubsection*{Detection Results on the VisA Dataset}

\begin{table*}[t]
\centering
\caption{\revred{Image-level anomaly detection performance (\%, I-AUROC / I-AP) on VisA under the OTFCL setting.}}
\label{tab:table2}
\begin{adjustbox}{width=\textwidth, center}
\begin{tabular}{l|cccccccccccc|c}
\toprule
Method & Candle & Capsules & Cashew & Chewinggum & Fryum & Macaroni1 & Macaroni2 & Pcb1 & Pcb2 & Pcb3 & Pcb4 & Pipe\_fryum & Average \\
\midrule
ER
& 52.1/9.4 & 64.5/24.0 & 57.0/22.9 & 42.2/14.4
& 77.1/48.6 & 52.6/12.0 & 53.4/10.8 & 67.9/18.1
& 71.5/18.6 & 63.4/21.7 & 48.9/8.5 & 53.5/24.1
& 58.7/19.4 \\

DualPrompt
& 61.8/12.6 & 48.3/14.3 & 55.1/24.9 & 74.6/58.4
& 80.8/49.5 & 50.7/11.9 & 63.3/16.0 & 57.9/13.5
& 52.0/14.3 & 46.3/9.2 & 70.7/25.6 & 41.9/18.7
& 58.6/21.7 \\

UniAD
& 26.7/5.8 & 56.9/18.7 & 40.4/13.7 & 46.7/15.9
& 48.2/17.3 & 47.3/8.6 & 41.2/7.5 & 41.6/7.3
& 45.4/8.3 & 36.3/6.8 & 64.2/19.4 & 38.7/15.5
& 44.5/12.1 \\

L2P
& 79.3/39.8 & 57.0/21.9 & 62.4/30.7 & 83.1/64.9
& 65.0/30.5 & 75.5/25.8 & 58.7/13.0 & 50.9/13.9
& 56.8/26.0 & 43.5/8.1 & 71.8/21.5 & 40.0/16.6
& 62.0/26.0 \\

MVP
& \textbf{83.9}/39.2 & 60.4/19.5 & 28.2/14.5 & 76.0/49.8
& 64.4/29.4 & 67.5/22.2 & 58.6/33.9 & 80.4/33.9
& 55.4/14.8 & 43.3/7.6 & 63.0/24.6 & 68.1/35.8
& 62.4/25.2 \\

FCA*
& 18.5/5.3 & 30.1/9.7 & 82.7/57.2 & 29.6/17.8
& 27.3/11.0 & \textbf{80.7}/31.9 & 43.9/7.7 & 46.6/8.9
& 78.1/\textbf{56.4} & 61.3/16.6 & 18.6/5.3 & 56.3/24.4
& 50.4/22.0 \\

DCM
& 73.0/27.0 & 51.2/18.5 & \textbf{90.9}/71.3 & 60.3/22.7
& 21.2/10.2 & 19.0/5.3 & 67.5/16.7 & 69.7/19.9
& \textbf{79.7}/49.0 & 38.2/7.0 & 12.1/5.0 & 34.2/12.6
& 51.4/22.1 \\

DYSON
& 50.0/9.1 & 50.0/14.3 & 50.0/16.7 & 50.0/16.6
& 50.0/16.7 & 50.0/9.1 & 50.0/9.1 & 50.0/9.1
& 50.0/9.1 & 50.0/9.0 & 50.0/9.1 & 50.0/16.7
& 50.0/12.0 \\

DYSON*
& 69.9/23.3 & 59.8/20.3 & 87.3/74.0
& \textbf{91.5}/\textbf{88.0}
& 74.5/56.6 & 61.3/20.2 & 69.1/21.5 & 58.9/18.4
& 55.6/11.7 & 66.0/17.9 & 83.9/49.1 & 91.0/80.0
& 72.4/40.1 \\

Online-LoRA
& 54.3/11.9 & 31.8/9.8 & 66.7/34.1 & 25.4/11.3
& 62.6/32.6 & 73.6/21.3 & 55.8/18.0 & 80.9/15.9
& 64.5/16.5 & 57.7/7.8 & 43.8/15.9 & 39.5/15.9
& 54.7/18.4 \\

\revred{IUF}
& \revred{69.46/40.92}
& \revred{54.37/64.33}
& \revred{60.84/60.12}
& \revred{58.28/60.91}
& \revred{51.82/67.45}
& \revred{50.69/51.31}
& \revred{58.42/45.37}
& \revred{62.27/43.17}
& \revred{53.73/50.14}
& \revred{50.24/48.38}
& \revred{51.79/51.42}
& \revred{60.10/76.85}
& \revred{56.83/55.03} \\

\revred{CDAD}
& \revred{63.96/\textbf{67.76}}
& \revred{57.45/59.20}
& \revred{64.66/\textbf{77.39}}
& \revred{51.38/66.08}
& \revred{55.56/73.78}
& \revred{54.83/50.40}
& \revred{52.30/52.77}
& \revred{59.08/58.59}
& \revred{57.96/49.30}
& \revred{57.04/\textbf{60.49}}
& \revred{52.25/54.62}
& \revred{73.90/\textbf{84.81}}
& \revred{58.36/62.93} \\

\revred{UCAD}
& \revred{57.44/58.29}
& \revred{\textbf{67.53}/\textbf{75.64}}
& \revred{51.50/67.51}
& \revred{61.98/78.35}
& \revred{54.74/\textbf{75.84}}
& \revred{42.01/53.93}
& \revred{58.32/\textbf{61.96}}
& \revred{63.10/\textbf{66.15}}
& \revred{51.67/54.88}
& \revred{52.36/54.93}
& \revred{48.32/50.35}
& \revred{47.70/70.95}
& \revred{54.72/\textbf{64.06}} \\

\revred{IB-IUMAD}
& \revred{67.61/42.01}
& \revred{56.08/66.73}
& \revred{55.80/63.25}
& \revred{57.86/62.06}
& \revred{50.66/68.26}
& \revred{55.11/\textbf{54.74}}
& \revred{58.56/44.81}
& \revred{54.99/47.18}
& \revred{51.40/48.48}
& \revred{52.09/50.91}
& \revred{60.35/\textbf{60.77}}
& \revred{60.64/76.77}
& \revred{56.76/57.16} \\

Ours
& \revred{75.6/35.0}
& \revred{62.4/22.4}
& 85.7/71.4
& \revred{90.2/86.6}
& \revred{\textbf{81.1}/67.8}
& 67.4/23.3
& \revred{\textbf{74.2}/27.9}
& \revred{\textbf{85.4}/37.2}
& 66.4/18.0
& \revred{\textbf{66.4}/21.0}
& \revred{\textbf{84.7}/51.5}
& \revred{\textbf{92.6}/81.1}
& \revred{\textbf{77.7}/45.3} \\
\bottomrule
\end{tabular}
\end{adjustbox}

\smallskip
\raggedright
\footnotesize
$^*$FCA$^*$ and DYSON$^*$ are NC-based continual learning methods; synthetic anomaly samples were used to meet their two-class training requirement.
\end{table*}

\revred{The image-level detection results on the VisA dataset are summarized in Table~\ref{tab:table2}. Compared with MVTec AD, VisA contains more complex backgrounds and more diverse anomaly patterns, making task-free continual anomaly detection more challenging. NC-TFAD achieves the highest average I-AUROC of 77.7\%, outperforming the second-best DYSON* by 5.3 percentage points, and shows strong performance across several challenging categories, including Fryum, Macaroni2, Pcb1, Pcb3, Pcb4, and Pipe\_fryum. However, its average I-AP of 45.3\% is lower than that of several industrial anomaly detection baselines, with UCAD achieving the highest average I-AP of 64.06\%. This discrepancy is likely related to the severe class imbalance in VisA: while AUROC is relatively insensitive to class prevalence and primarily reflects overall ranking discrimination, AP is more sensitive to false positives among high-confidence predictions. Therefore, the VisA results demonstrate that NC-TFAD provides strong overall normal--anomaly discrimination, but its precision--recall behavior under highly imbalanced streams remains a limitation. Improving anomaly-score calibration and suppressing high-confidence false positives constitute promising directions for future work.}

\subsubsection*{Localization Results}

\revorange{
Most task-free and incremental learning methods considered in the image-level evaluation are designed for image-level prediction and do not natively produce dense anomaly maps. Introducing additional localization modules would alter their original architectures and make the resulting comparison dependent on the specific localization design. Therefore, the pixel-level evaluation is conducted on methods that inherently support anomaly localization, including UniAD, IUF, UCAD, CDAD, and IB-IUMAD. 
}

\begin{figure*}[!t]
\centering
\includegraphics[width=\textwidth]{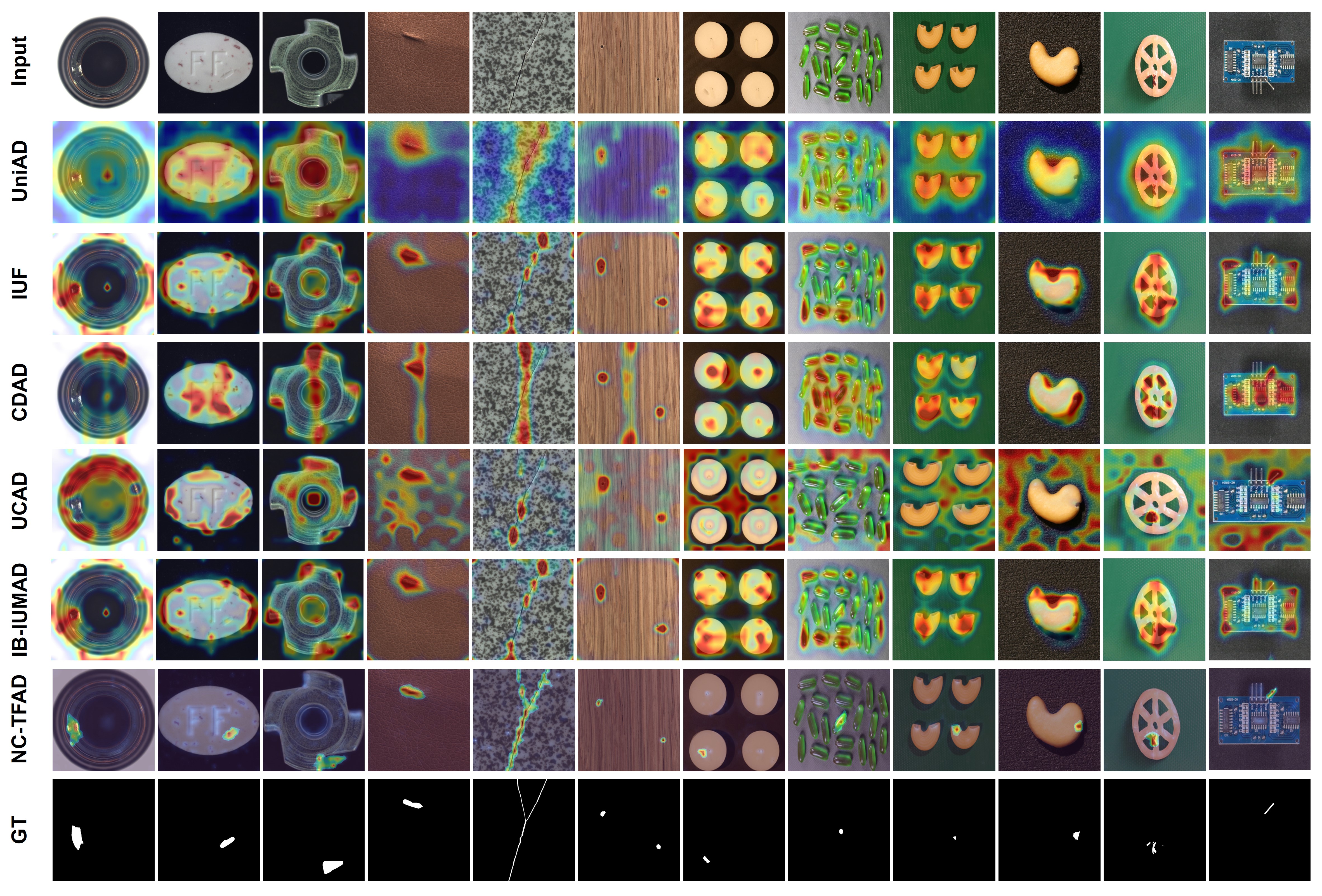}
\caption{Qualitative visualizations of anomaly localization on MVTec AD and VisA. The first six columns show results on MVTec AD, and the last six columns show results on VisA.}
\label{fig3}
\end{figure*}

\begin{table*}[!t]
\centering
\caption{\revorange{Pixel-level anomaly localization performance (\%) on MVTec AD and VisA under the continual learning setting. Results are reported in terms of P-AUROC, P-AUPRO, and P-AP.}}
\label{tab:localization_results}
\begin{adjustbox}{width=0.82\textwidth,center}
\begin{tabular}{lccc|ccc}
\toprule
\multirow{2}{*}{Method}
& \multicolumn{3}{c|}{MVTec AD}
& \multicolumn{3}{c}{VisA} \\
\cmidrule(lr){2-4}\cmidrule(lr){5-7}
& P-AUROC $\uparrow$
&\revorange{P-AUPRO $\uparrow$}
& P-AP $\uparrow$
& P-AUROC $\uparrow$
&\revorange{P-AUPRO $\uparrow$}
& P-AP $\uparrow$ \\
\midrule

\revorange{IUF}
&\revorange{77.38}
&\revorange{55.02}
&\revorange{14.76}
&\revorange{85.35}
&\revorange{53.30}
&\revorange{3.34} \\

\revorange{CDAD}
&\revorange{75.53}
&\revorange{47.97}
&\revorange{8.79}
&\revorange{85.98}
&\revorange{56.04}
&\revorange{\textbf{5.54}} \\

\revorange{UCAD}
&\revorange{73.56}
&\revorange{51.92}
&\revorange{22.38}
&\revorange{42.59}
&\revorange{6.16}
&\revorange{1.13} \\

\revorange{IB-IUMAD}
&\revorange{77.45}
&\revorange{56.04}
&\revorange{14.78}
&\revorange{85.27}
&\revorange{53.62}
&\revorange{3.49} \\

UniAD
& 81.70
&\revorange{50.70}
& 17.20
& 84.90
&\revorange{53.80}
& 0.60 \\

\textbf{NC-TFAD}
& \textbf{87.54}
&\revorange{\textbf{73.49}}
& \textbf{34.93}
& \textbf{88.91}
&\revorange{\textbf{69.83}}
& 2.85 \\

\bottomrule
\end{tabular}
\end{adjustbox}
\end{table*}

\revorange{
Table~\ref{tab:localization_results} reports the pixel-level localization performance on MVTec AD and VisA in terms of P-AUROC, P-AUPRO, and P-AP. On MVTec AD, NC-TFAD achieves the best average performance across all three metrics, reaching 87.54\% P-AUROC, 73.49\% P-AUPRO, and 34.93\% P-AP. These results demonstrate that the proposed prototype-calibrated localization branch effectively identifies anomalous pixels while preserving coherent defect regions. On VisA, NC-TFAD further achieves the highest P-AUROC and P-AUPRO, with 88.91\% and 69.83\%, respectively, demonstrating strong pixel-level discrimination and region-level localization under more complex anomaly patterns. Its P-AP remains comparatively low, which is mainly attributable to the small and sparse defects in VisA, where a limited number of false-positive pixels can substantially affect precision under severe pixel-level class imbalance. This limitation and potential improvements in fine-grained anomaly-score calibration are further discussed in the Limitations section.
}

\revorange{
The qualitative results in Fig.~\ref{fig3} are consistent with these quantitative observations. The prototype-calibrated deviation responses are generally concentrated around defect-related regions, while the attention prior and connected-region filtering reduce isolated background responses and preserve spatially coherent anomaly regions. Nevertheless, localization remains challenging for small or weakly contrasted defects, particularly on VisA. These results suggest that pixel-level localization remains a challenging auxiliary task under the OTFCL setting.
}

\subsection{Ablation Study}

\subsubsection*{Module Effectiveness}
Ablation results for different module configurations are presented in Table~\ref{tab:table5}. When both NC regularization and FNCC loss are disabled, the model performance is substantially limited. This effect is particularly pronounced on the VisA dataset, indicating that basic feature learning alone is insufficient to handle complex distribution shifts in a task-free data stream.

\begin{table}[!t]
\centering
\caption{Ablation study of model components}
\label{tab:table5}
\begin{tabular}{l l c c c c}
\toprule
\multirow{2}{*}{Regularizer} &\multirow{2}{*}{ FNCC Loss} & \multicolumn{2}{c}{MvTec} & \multicolumn{2}{c}{VisA} \\
\cmidrule(lr){3-4} \cmidrule(lr){5-6}
 & & I-AUROC & I-AP & I-AUROC & I-AP \\
\midrule
$\times$ & $\times$ & 75.5 & 88.6 & 72.0 & 39.7 \\
$\surd$ & $\times$ & 81.2 & 91.5 & 74.0 & 41.4 \\
$\times$ & $\surd$ & 80.2 & 90.7 & 77.4 & 44.7 \\
$\surd$ & $\surd$ & \textbf{82.1} & \textbf{91.8} & \textbf{77.7} & \textbf{45.3} \\
\bottomrule
\end{tabular}
\end{table}

When only NC-guided regularization is applied, the performance on MVTec AD improves noticeably. Specifically, I-AUROC increases from 75.5\% to 81.2\%, while I-AP rises to 91.5\%. This result confirms that NC-based regularization effectively alleviates representation drift and reinforces the geometric decision structure. When only the FNCC loss is introduced, performance gains on VisA are more prominent, highlighting the importance of hard-sample-aware contrastive optimization in challenging continual settings.

When both components are enabled simultaneously, the model achieves the best overall performance on both datasets. This observation validates the strong complementarity between geometric regularization and focal contrastive learning.

\subsubsection*{Effect of Inter- and Intra-Class Regularization}
The impact of the inter-class and intra-class regularization terms on NC-TFAD is examined in Table~\ref{tab:table6}. When both constraints are removed, the model performance drops markedly on both datasets, indicating that the decision boundary becomes unstable under continual distribution shifts.

\begin{table}[!t]
\centering
\caption{Ablation study on inter- and intra-class regularization}
\label{tab:table6}
\begin{tabular}{l l c c c c}
\toprule
\multicolumn{2}{c}{Regularizer} & \multicolumn{2}{c}{MVTec} & \multicolumn{2}{c}{VisA} \\
\cmidrule(lr){1-2} \cmidrule(lr){3-4} \cmidrule(lr){5-6}
 $\lambda_{\text{inter}}$ & $\lambda_{\text{intra}}$ & I-AUROC & I-AP & I-AUROC & I-AP \\
\midrule
0.0 & 0.0 & 75.5 & 88.6 & 72.0 & 39.7 \\
0.1 & 0.05 & 79.2 & 89.7 & 75.6 & 43.2 \\
0.1 & 0.2 & 79.3 & 87.3 & 75.4 & 43.2 \\
0.2 & 0.1 & 77.6 & 89.5 & 74.0 & 41.4 \\
0.3 & 0.1 & 78.4 & 89.6 & \textbf{76.2} & 43.9 \\
0.4 & 0.15 & 78.1 & 87.1 & 75.8 & 43.2 \\
0.5 & 0.2 & 78.9 & 90.0 & 75.9 & 43.1 \\
0.9 & 0.5 & \textbf{81.2} & \textbf{91.5} & 76.1 & \textbf{43.9} \\
\bottomrule
\end{tabular}
\end{table}

As the strengths of the inter-class and intra-class regularizers are gradually increased, the performance exhibits a consistent upward trend. On MVTec AD, the model reaches its best performance when the regularization strength is further increased. With $\lambda_{\text{inter}} = 0.9$ and $\lambda_{\text{intra}} = 0.5$, I-AUROC and I-AP are improved to 81.2\% and 91.5\%, respectively. On the VisA dataset, the best results are obtained in a medium-to-high regularization regime, where I-AUROC reaches up to 76.2\% ($\lambda_{\text{inter}} = 0.3$) and I-AP 43.9\% ($\lambda_{\text{inter}}= 0.3$ or $0.9$). Compared with MVTec AD, the more complex backgrounds and diverse anomaly patterns in VisA make the performance more sensitive to the choice of regularization weights.

\subsubsection*{Sensitivity to the Contrastive Temperature Parameter}
Table~\ref{tab:table7} reports the sensitivity analysis of the FNCC loss with respect to the temperature parameter $\tau$. The temperature controls the smoothness of the similarity distribution and thus affects the sharpness of the decision boundaries between samples.

Without NC-guided regularization, the model is highly sensitive to variations in $\tau$. As $\tau$ increases from 0.05 to 0.20, the performance on both datasets improves significantly. For example, on VisA, I-AUROC rises from 72.0\% to 77.4\% and I-AP from 39.7\% to 44.7\%. However, when $\tau$ continues to increase (e.g., $\tau \geq 0.30$), the performance begins to degrade. This behavior indicates that an excessively large temperature weakens discriminative learning on hard samples.

After jointly incorporating the NC regularization and FNCC, the model becomes substantially more stable across a wide range of temperature values. For $0.1 \leq \tau \leq 0.3$, performance remains consistently high on both datasets and exceeds that of the setting without regularization. Among these configurations, $\tau = 0.1$ yields the best results, achieving 82.1\% / 91.8\% on MVTec AD and 77.7\% / 45.3\% on VisA in terms of I-AUROC / I-AP.

\begin{table}[t]
\centering
\caption{Ablation study on the sensitivity of the contrastive temperature $\tau$}
\label{tab:table7}
\begin{tabular}{@{}ccccccc@{}}
\toprule
\multirow{2}{*}{FNCC loss}& \multirow{2}{*}{Reg.} & \multirow{2}{*}{Temp.($\tau$)}  & \multicolumn{2}{c}{MVTec} & \multicolumn{2}{c}{VisA} \\
\cmidrule(lr){4-5} \cmidrule(lr){6-7}
& & & I-AUROC & I-AP & I-AUROC & I-AP \\
\midrule
$\surd$ & $\times$ & 0.05 & 75.5 & 88.6 & 72.0 & 39.7 \\
$\surd$ & $\times$ & 0.1  & 80.2 & 90.7 & 77.1 & 44.4 \\
$\surd$ & $\times$ & 0.2  & 80.4 & 90.5 & 77.4 & 44.7 \\
$\surd$ & $\times$ & 0.3  & 79.9 & 90.4 & 76.4 & 43.2 \\
$\surd$ & $\times$ & 0.5  & 79.4 & 90.1 & 75.9 & 42.7 \\
\midrule
$\surd$ & $\surd$ & 0.05 & 80.8 & 90.9 & 75.9 & 43.3 \\
$\surd$ & $\surd$ & 0.1  & \textbf{82.1} & \textbf{91.8} & \textbf{77.7} & \textbf{45.3} \\
$\surd$ & $\surd$ & 0.2  & 82.0 & 91.6 & 77.2 & 44.5 \\
$\surd$ & $\surd$ & 0.3  & 81.8 & 91.4 & 77.1 & 44.5 \\
$\surd$ & $\surd$ & 0.5  & 81.4 & 91.2 & 76.8 & 44.1 \\
\bottomrule
\end{tabular}
\end{table}

\begin{table*}[t]
\centering
\caption{\rev{Sensitivity analysis of minibatch composition on MVTec AD and VisA. Results are reported as mean ± standard deviation over three random seeds.}}
\label{tab:batch_composition}
\resizebox{\textwidth}{!}{
\begin{tabular}{lcccccccc}
\toprule
\multirow{2}{*}{Setting} & \multicolumn{4}{c}{MVTec AD} & \multicolumn{4}{c}{VisA} \\
\cmidrule(lr){2-5} \cmidrule(lr){6-9}
& I-AUROC$\uparrow$ & I-AP$\uparrow$ & P-AUROC$\uparrow$ & P-AP$\uparrow$ & I-AUROC$\uparrow$ & I-AP$\uparrow$ & P-AUROC$\uparrow$ & P-AP$\uparrow$ \\
\midrule
Random & $\mathbf{80.01{\pm}0.74}$ & $90.57{\pm}0.53$ & $87.36{\pm}0.27$ & $30.77{\pm}0.18$ & $77.04{\pm}0.92$ & $47.52{\pm}1.35$ & $\mathbf{89.02{\pm}0.17}$ & $\mathbf{2.14{\pm}0.09}$ \\
Homogeneous & $79.54{\pm}1.61$ & $90.11{\pm}0.94$ & $87.25{\pm}0.21$ & $30.27{\pm}0.09$ & $76.26{\pm}1.05$ & $44.13{\pm}1.74$ & $88.69{\pm}0.18$ & $1.82{\pm}0.15$ \\
Balanced & $79.62{\pm}0.95$ & $\mathbf{90.69{\pm}0.64}$ & $\mathbf{87.60{\pm}0.14}$ & $\mathbf{30.85{\pm}0.33}$ & $\mathbf{77.89{\pm}1.27}$ & $\mathbf{48.03{\pm}1.70}$ & $88.93{\pm}0.15$ & $2.12{\pm}0.15$ \\
\bottomrule
\end{tabular}}
\end{table*}

\subsubsection*{Effect of Minibatch Composition}

\rev{To assess sensitivity to minibatch composition, we consider three stream arrangements: Random, which globally shuffles samples; Homogeneous, which groups samples by product category; and Balanced, which uses round-robin sampling to increase category diversity within each batch. All settings share the same samples, batch size, optimization configuration, and single-pass protocol. Category identities are used only to construct this diagnostic ablation and are never provided to NC-TFAD during training or inference.}

\rev{As shown in Table~\ref{tab:batch_composition}, NC-TFAD is generally robust to different minibatch compositions. The three settings yield comparable results on MVTec AD, while on VisA the Homogeneous setting is consistently weaker than Random and Balanced. This suggests that the shared normal--anomaly geometry does not depend on category-homogeneous batches, although greater cross-category exposure can benefit more composition-sensitive streams.}

\subsubsection*{Temporal Retention Analysis}

\rev{We compare four variants under the same task-free continual stream: \emph{base} retains the fixed ETF alignment but excludes NC-guided regularization and FNCC, \emph{+ Reg.} and \emph{+ FNCC} introduce the two objectives individually, and NC-TFAD combines both. Final I-AUROC and Final I-AP evaluate overall image-level detection at the end of the stream, while Old-category I-AUROC evaluates the final model only on categories observed before the last stream segment to measure retained detection capability. Category identities are used only for offline evaluation.}

\begin{table}[!t]
\centering
\caption{\rev{Temporal retention analysis on MVTec AD and VisA. Results are reported as mean$\pm$standard deviation over three random seeds.}}
\label{tab:temporal_retention}
\resizebox{\columnwidth}{!}{
\begin{tabular}{llccc}
\toprule
Dataset & Variant & Final I-AUROC$\uparrow$ & Final I-AP$\uparrow$ & Old-category I-AUROC$\uparrow$ \\
\midrule
\multirow{4}{*}{MVTec AD}
& base & $76.11{\pm}0.40$ & $88.90{\pm}0.13$ & $77.41{\pm}0.09$ \\
& + Reg. & $79.22{\pm}1.92$ & $90.25{\pm}0.81$ & $80.25{\pm}1.62$ \\
& + FNCC & $79.12{\pm}0.78$ & $90.01{\pm}0.44$ & $79.38{\pm}0.72$ \\
& NC-TFAD & $\mathbf{79.76{\pm}1.40}$ & $\mathbf{90.34{\pm}0.72}$ & $\mathbf{80.60{\pm}1.18}$ \\
\midrule
\multirow{4}{*}{VisA}
& base & $71.16{\pm}0.52$ & $38.41{\pm}0.76$ & $72.28{\pm}0.38$ \\
& + Reg. & $74.42{\pm}1.25$ & $41.76{\pm}1.59$ & $74.58{\pm}1.43$ \\
& + FNCC & $75.69{\pm}0.26$ & $42.66{\pm}0.35$ & $75.44{\pm}0.41$ \\
& NC-TFAD & $\mathbf{76.33{\pm}0.08}$ & $\mathbf{43.70{\pm}0.26}$ & $\mathbf{77.03{\pm}0.39}$ \\
\bottomrule
\end{tabular}}
\end{table}

\rev{As shown in Table~\ref{tab:temporal_retention}, both NC-guided regularization and FNCC improve final and old-category performance over the base model, while their combination achieves the strongest results across both datasets. This indicates that the proposed objectives improve retention of earlier distributions while maintaining strong end-of-stream detection performance.}

\subsubsection*{\revorange{Representation Geometry Dynamics}}

\begin{figure*}[!t]
\centering
\includegraphics[width=0.8\textwidth]{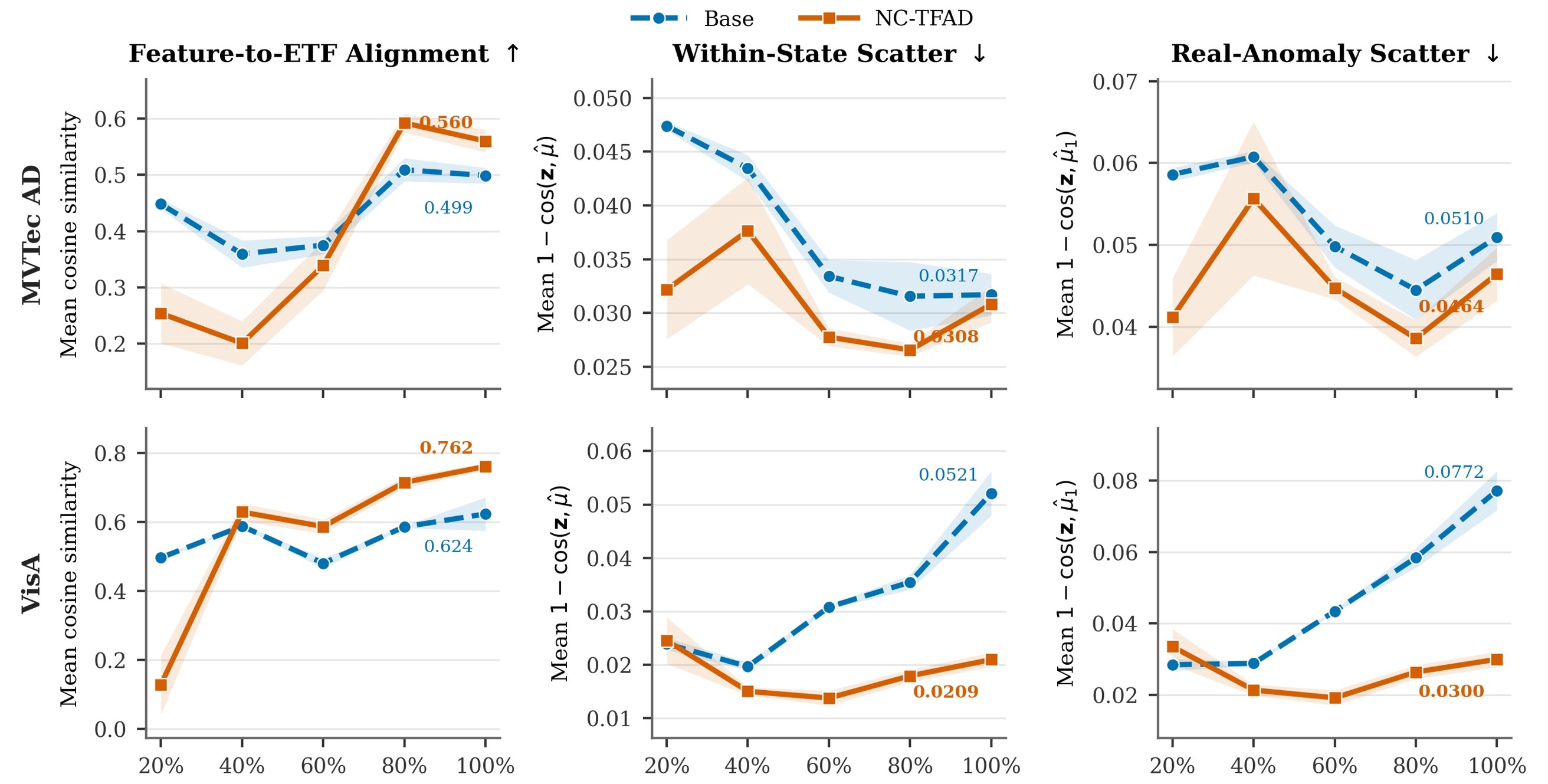}
\caption{\revorange{Representation-geometry dynamics over the continual stream. Curves and shaded regions denote the mean and one standard deviation over three random seeds, respectively. Feature-to-ETF Alignment measures the cosine similarity between projected features and their assigned fixed ETF prototypes. Within-State Scatter measures the dispersion of normal and anomalous representations around their corresponding mean embeddings, whereas Real-Anomaly Scatter evaluates the compactness of real anomalous representations only. Lower scatter indicates more compact representations.}}
\label{fig:geometry_dynamics}
\end{figure*}

\revorange{To examine how the representation geometry evolves during continual learning, we save read-only checkpoints at 20\%, 40\%, 60\%, 80\%, and 100\% of the single-pass stream and evaluate them offline after training. This procedure neither alters optimization nor revisits any training sample. At each checkpoint, we evaluate three quantities shown in Fig.~\ref{fig:geometry_dynamics}. Feature-to-ETF Alignment is the mean cosine similarity between each projected embedding and the fixed ETF prototype associated with its anomaly state. Within-State Scatter measures the cosine-distance dispersion around the corresponding mean embedding, computed separately for each product category and anomaly state and then macro-averaged. Real-Anomaly Scatter uses the same measure but is restricted to real anomalous samples. Product-category identities and real anomaly labels are used only for these offline statistics and are never provided to the model during continual optimization.}

\revorange{As shown in Fig.~\ref{fig:geometry_dynamics}, NC-TFAD consistently strengthens feature-to-ETF alignment while producing more compact representations than the base model, with particularly pronounced improvements on VisA and for real anomalous samples. These trends support the intended role of NC-guided regularization and FNCC in preserving prototype anchoring and representation compactness as the data distribution evolves.}

\subsubsection*{Effect of Different Backbones}

\revred{To assess the sensitivity of NC-TFAD to the pretrained feature extractor, we evaluate the framework with DINO ViT-S/8, ViT-S/16, ResNet-50, and EfficientNet-B4 under the same OTFCL protocol, with all backbone parameters kept frozen throughout training. As shown in Table~\ref{tab:backbone_ablation}, NC-TFAD remains effective across both Transformer- and convolution-based architectures. Moreover, comparisons under matched backbone settings consistently show stronger I-AUROC than the corresponding representative baselines. These results demonstrate that the effectiveness of the proposed NC-guided framework is not restricted to the default DINO ViT-S/8 backbone and generalizes across heterogeneous feature extractors.}

\begin{table}[!t]
\centering
\caption{\revred{Effect of different frozen backbones on NC-TFAD and architecture-related comparison with representative methods \revorange{under the OTFCL protocol}. Results are reported as I-AUROC / I-AP (\%).}}
\label{tab:backbone_ablation}
\resizebox{\columnwidth}{!}{
\begin{tabular}{llcc}
\toprule
Backbone & Method & MVTec AD & VisA \\
\midrule
\multicolumn{4}{c}{\textit{NC-TFAD with different frozen backbones}} \\
\midrule

ViT-S/16
& NC-TFAD & 72.18 / 87.04 & 63.83 / 32.09 \\

ResNet-50
& NC-TFAD & 76.45 / 87.81 & 63.54 / 28.17 \\

EfficientNet-B4
& NC-TFAD & 77.74 / 89.53 & 69.10 / 35.90 \\

DINO ViT-S/8
& \textbf{NC-TFAD} & \textbf{82.10 / 91.80} & \textbf{77.70 / 45.30} \\

\midrule
\multicolumn{4}{c}{\textit{Architecture-related comparison with representative methods}} \\
\midrule

\multirow{2}{*}{DINO ViT-S/8}
& DYSON$^{*}$ & 70.10 / 86.30 & 72.40 / 40.10 \\
& \textbf{NC-TFAD} & \textbf{82.10 / 91.80} & \textbf{77.70 / 45.30} \\

\addlinespace

\multirow{2}{*}{ViT-S/16}
& Online-LoRA & 61.40 / 81.30 & 54.70 / 18.40 \\
& \textbf{NC-TFAD} & \textbf{72.18 / 87.04} & \textbf{63.83 / 32.09} \\

\addlinespace

\multirow{3}{*}{ResNet-family}
& CDAD (ResNet-50) & 60.39 / 78.79 & 58.36 / 62.93 \\
& UCAD (WideResNet-50) & 72.07 / 87.38 & 54.72 / \textbf{64.07} \\
& \textbf{NC-TFAD} (ResNet-50) & \textbf{76.45} / \textbf{87.81} & \textbf{63.54} / 28.17 \\

\addlinespace

\multirow{4}{*}{EfficientNet-B4}
& UniAD & 63.10 / 82.00 & 44.50 / 12.10 \\
& IUF & 70.87 / 82.79 & 56.83 / 55.03 \\
& IB-IUMAD & 71.17 / 82.85 & 56.76 / \textbf{57.16} \\
& \textbf{NC-TFAD} & \textbf{77.74} / \textbf{89.53} & \textbf{69.10} / 35.90 \\

\bottomrule
\end{tabular}}
\vspace{1mm}

\raggedright
\footnotesize
$^{*}$DYSON$^{*}$ denotes the variant trained with synthetic anomaly samples to satisfy its two-class training requirement. \revred{CDAD, UCAD, IUF, and IB-IUMAD are reproduced under the same OTFCL stream used in the SOTA comparison.}
\end{table}

\subsubsection*{Effect of Synthetic Anomaly Strategies}

\revred{To examine the influence of synthetic anomaly construction, we compare several synthesis variants on MVTec AD under the same task-free continual learning protocol and fixed seed. As summarized in Table~\ref{tab:synthetic_ablation}, all variants are trained with synthetic anomalies but evaluated exclusively on real anomalous test images.}

\begin{table}[!t]
\centering
\caption{\revred{Ablation study of synthetic anomaly strategies on MVTec AD. All variants follow the same task-free continual learning protocol and are evaluated on real anomalous test images. The full strategy denotes the foreground-constrained local mixed synthesis used in NC-TFAD.}}
\label{tab:synthetic_ablation}
\resizebox{\columnwidth}{!}{
\begin{tabular}{lccccc}
\toprule
\textbf{Synthetic Strategy} & \textbf{I-AUROC} & \textbf{I-AP} & \textbf{P-AUROC} & \textbf{P-AP} & \textbf{P-AUPRO} \\
\midrule
\revred{Foreground + appearance only} & \revred{63.77} & \revred{83.96} & \revred{87.07} & \revred{28.57} & \revred{70.96} \\
\revred{Foreground + local blur only} & \revred{56.61} & \revred{80.16} & \revred{87.31} & \revred{28.62} & \revred{70.45} \\
\revred{Foreground + local rearrangement only} & \revred{79.18} & \revred{90.43} & \revred{86.95} & \revred{29.29} & \revred{72.20} \\
\revred{Global perturbation + mixed} & \revred{53.01} & \revred{76.75} & \revred{86.45} & \revred{27.30} & \revred{70.70} \\
\revred{Random local mask + mixed} & \revred{77.25} & \revred{89.82} & \revred{87.47} & \revred{29.00} & \revred{71.63} \\
\revred{\textbf{Full strategy}} & \revred{\textbf{82.10}} & \revred{\textbf{91.80}} & \revred{\textbf{87.54}} & \revred{\textbf{34.93}} & \revred{\textbf{73.49}} \\
\bottomrule
\end{tabular}}
\end{table}

\revred{As shown in Table~\ref{tab:synthetic_ablation}, local structural perturbations are more effective than simple appearance or blur transformations, while global perturbations substantially degrade image-level detection. The full strategy consistently provides the strongest overall performance, indicating that combining spatially plausible local regions with diverse perturbation types yields more effective auxiliary anchors for both image-level geometric learning and anomaly localization.}

\subsection{Limitations}

\revred{Despite the encouraging results, NC-TFAD has several limitations. First, synthetic anomalies are used as auxiliary geometric anchors because real defects are unavailable during training, and their effectiveness therefore depends on whether the generated perturbations provide sufficiently informative normal--anomaly cues; they cannot fully capture the appearance and structural diversity of real industrial defects. Second, although the proposed shared geometry is generally robust to different stream compositions, performance can still vary under more challenging data arrangements, particularly on VisA. Third, the current evaluation is limited to MVTec AD and VisA, which are constructed benchmark datasets rather than naturally collected long-term industrial streams, leaving the behavior of NC-TFAD under gradual, recurrent, or irregular real-world distribution shifts to be further investigated. Finally, the pixel-level comparison is restricted to methods that natively support anomaly localization, and the relatively low AP observed under severe class imbalance indicates that finer anomaly-score calibration remains an open issue. These limitations motivate future research on more adaptive anomaly synthesis, robust continual adaptation under realistic temporal shifts, and improved calibration and localization for highly imbalanced industrial data.}

\section{Conclusion}
This paper formulates industrial visual inspection as a task-free continual anomaly detection problem, where data streams evolve over time without observable task boundaries. \revorange{To address this setting, we propose NC-TFAD, a neural-collapse-inspired geometry-driven framework that establishes a category-agnostic normal--anomaly reference using fixed ETF prototypes. Building on this geometry, NC-guided inter- and intra-class regularization and the Focal Neural Collapse Contrastive loss jointly promote prototype alignment, representation compactness, and sample-level discriminability under non-stationary streams. A normal-patch-prototype-guided localization branch further combines calibrated patch deviations with a weak attention prior to enable pixel-level anomaly localization without pixel-level supervision.
}

\revorange{Extensive experiments and ablation studies on MVTec AD and VisA demonstrate that NC-TFAD achieves strong and competitive image-level anomaly detection performance and competitive pixel-level localization performance against representative localization-capable baselines under the task-free continual learning protocol. The results further show that the proposed geometric constraints improve representation stability and preserve discriminative capability as the data distribution evolves. Future work will investigate more adaptive synthetic anomaly generation, anomaly-score calibration under severe class imbalance, continual adaptation on naturally collected long-term industrial streams, and more general localization mechanisms across broader industrial scenarios.}

\end{document}